\documentclass[11pt]{article} 
\usepackage[paperheight=9.44in,paperwidth=6.69in,left=1.85cm,right=1.8cm,bottom=2.5cm,top=3cm,twoside]{geometry}

\usepackage{subfig} 

\usepackage[ruled,vlined,boxed]{algorithm2e}
\usepackage{authblk} 
\usepackage{fancyhdr} 
\usepackage{lastpage} 
\usepackage{textcomp} 
\usepackage{amsmath} 
\usepackage{amssymb} 
\usepackage{amsthm} 
\usepackage{graphicx} 
\usepackage{multirow} 
\usepackage{hyperref} 
\usepackage{longtable} 
\usepackage{float} 
\usepackage{enumitem} 
\usepackage[T1]{fontenc} 
\usepackage[utf8]{inputenc} 

\newcommand{\linia}{\noindent\rule{\linewidth}{0.25mm}\hrulefill} 

\SetAlFnt{\small}
\SetAlCapFnt{\small}
\SetAlCapNameFnt{\small}
\SetAlCapHSkip{0pt}
\IncMargin{-\parindent}

\usepackage{times}
\usepackage{titlesec}

\titleformat*{\section}{\large\bfseries}
\titleformat*{\subsection}{\normalsize\bfseries}

\fancypagestyle{firststyle}
{
\fancyhf{}
\setcounter{page}{57} 
\fancyfoot{}
\fancyhead{}
\fancyhead[CE]{\upshape Zeszyty Naukowe WWSI, No 27, Vol. 16, 2022, pp.\thepage-\pageref{LastPage} \newline DOI: 10.26348/znwwsi.27.\thepage}
\fancyhead[CO]{\upshape Zeszyty Naukowe WWSI, No 27, Vol. 16, 2022, pp. \thepage-\pageref{LastPage} \newline DOI: 10.26348/znwwsi.27.\thepage}
\fancyfoot[L,R]{\footnotesize\bfseries Manuscript received November 7, 2022}
\fancyfoot[LE,RO]{\thepage}
}

\title{\large \bfseries On the Analysis of Correlation \\Between Nominal Data and Numerical Data} 
\author{\normalsize Zenon Gniazdowski\thanks{E-mail: zgniazdowski@wwsi.edu.pl}}
\affil{\normalsize Warsaw School of Computer Science}

\date{\vspace{-5ex}}
\providecommand{\keywords}[1]{\textbf{\textit{Keywords ---}} #1}

\begin{document}
	
	\maketitle 
	\thispagestyle{firststyle} 
	
	\linia
	
	\begin{abstract}\label{abstract}
		\noindent The article investigates the possibility of measuring the strength of a linear correlation relationship between nominal data and numerical data. Correlation coefficients for variables coded with real numbers as well as for variables coded with complex numbers were studied. For variables coded with real numbers, unambiguous measures of real linear correlation were obtained. In the case of complex coding, it has been observed that the obtained complex correlation coefficients change with the permutation of the phases in the complex numbers used to code classes of elements with equal cardinalities. It was found that a necessary condition for linear correlation is the possibility of linear ordering of a set with data. Since linear order is not possible in the set of complex numbers, complex correlation coefficients cannot be used as a measure of linear correlation. In the event of such a situation, a substitute action was suggested that would prevent equal cardinality of classes of identical elements contained in the set with nominal data. This action would consist in the correction of data, analogous to the correction during preprocessing or cleaning of data containing missing or outlier values.
	\end{abstract}
	\label{keywords}
	
	\section{Introduction}
	In classical statistics, the $\chi^2$ test is used to test the correlation between nominal variables. A contingency table is created for nominal data. On its basis, the  $\chi^2$  statistic can be used to assess the significance of the correlation between two nominal variables. The V-Cramer coefficient, also estimated using  $\chi^2$  statistics, can be used to measure the strength of this relationship \cite{Blalock1960}. On the other hand, for variables measured in at least an ordinal scale, the Pearson correlation coefficient or the Spearman rank correlation coefficient \cite{Blalock1960}\cite{Francuz2007} are examined.
	
	In \cite{Gniazdowski2015} a different approach to the problem was proposed. In this approach, nominal data is given a numerical interpretation. Since a random variable measured on a nominal scale takes k different values, each of the k subsets of identical values will be called a class later in this article. Depending on the cardinality of different classes, the elements of a given class will be coded with real numbers or complex numbers. If there are no classes with equal cardinality, then a given class will be assigned a real number which is a function of the cardinality of the class. If there are m classes of equal cardinality in the considered set of values, then each of these classes is assigned a complex number whose modulus is a function of the cardinality of a given class, and the phase is equal to the phase of one of the m roots of unity. The phases are chosen arbitrarily. However, it is required that the phases for each of the classes with identical cardinalities be different.
	
	The properties of numerical coding make it possible to cluster or classify nominal data using algorithms specific to numerical data \cite{Gniazdowski2015}. 
	It remains an open question whether the numerical coding of nominal data mentioned here can be used to assess the level of correlation between coded nominal  random variables. This article will attempt to answer this question.  For this purpose, the possibility of measuring the strength of the correlation between variables measured in the nominal scale and variables measured in the nominal scale or at least in the ordinal scale will be analyzed. The starting point for this analysis will be two observations:
	\begin{itemize}
		\item The correlation coefficient between two random variables has an interpretation of the cosine of the angle between the vectors containing the random components of these variables \cite{Gniazdowski2013}.
		\item Nominal data can be coded with real or complex numbers. Since in a real and complex vector space it is possible to define the scalar product and the Euclidean norm, it is also possible to calculate the cosine of the angle between the vectors \cite{Gniazdowski2015}.
	\end{itemize}
	The examples will show the influence of coding proposed in the paper \cite{Gniazdowski2015} on the possibility of identifying linear correlation between nominal variables. First, examples   of correlation analysis for nominal variables described by real numbers will be shown. Next , the possibility  of correlation analysis for nominal variables described by complex numbers will be examined and discussed.
	
	The analysis will be carried out on artificial data, specially prepared for the purposes of this work\footnote{To present the discussed problems, the author of this article uses artificially prepared data, because so far he has not encountered non-trivial nominal data sets that would contain classes of identical elements with equal cardinalities.}. Sample input data for the analysis will be provided in the form of a contingency table. A contingency table provides a compact representation of data consisting of many records. Since the aim of this article is to attempt to assess the possibility of measuring the strength of correlation between a variable measured on a nominal scale and a variable measured on a  nominal or stronger scale, the analyzed data should be selected in such a way that at least one variable can be interpreted as if it had been measured on a scale stronger than nominal, and therefore at least on an ordinal scale. For this reason, an additional restriction is imposed on the contingency table. It is required that after numerical coding of the nominal variables, the second variable may be treated as if it were measured on a scale stronger than the nominal scale:
	\begin{itemize}
		\item The first variable will be constructed in such a way that it can be coded using real numbers or complex numbers. In the case of coding with complex numbers, the first variable will contain at least two classes with identical cardinalities.  This will be manifested by the fact that at least two rows in the contingency table will have the same sums of elements.
		\item The second variable will always be constructed in such a way that it can only be coded with real numbers. This will only be possible if the variable does not contain classes with equal cardinalities. This will be manifested by the fact that in the columns of the contingency table the sums of elements will always be different in pairs. This approach will allow the second variable to be treated as if it were measured on a scale with a possible linear order relation, i.e. on a scale stronger than the nominal scale.
	\end{itemize}
	Since the second variable will be coded with real numbers with a well-defined linear order relation, the analyzed problem will become equivalent to the problem of examining the strength of the linear correlation relationship between a random variable measured in a nominal scale and a random variable measured in at least an ordinal scale. Thanks to this, the conclusions resulting from the study of correlations for two nominal variables will also be appropriate for the case of correlation between a variable measured in a nominal scale and a variable measured in one of the stronger scales: ordinal, interval or ratio.
	
	Finally, there is a comment on the accepted designations. In some mathematical formulas, a line will appear over the letter denoting the variable. If a given subsection concerns the geometric interpretation of correlations between random variables, the line will be placed over the capital letter and will represent the average value of the random variable. On the other hand, when a subsection deals with the definition of a scalar product for vectors containing complex numbers, or concerns correlation for complex random variables, then the line will be over lowercase letters and will indicate the conjugate of a complex number.
	\section{Preliminaries}
	The preliminaries will present some ideas to which the author will refer later in this article. 
	Here, the following concepts will be introduced:  binary relation, measurement scales, statistics $\chi^2$, strength of the correlation relationship for nominal data, numerical coding of nominal data, geometric interpretation of a correlation, as well as the least squares complex method.
	
	\begin{table}[t]
		\centering
		\caption{ Properties of a binary relation }\label{tab1}
		\fontsize{9}{13}\selectfont{
			\begin{tabular}{ c||c }
				The name of property&The essence of a given property\\ \hline\hline
				Reflexivity&$\forall x\in X:\;x\rho x$ \\\hline
				Irreflexivity&$\forall x\in X:\;\neg(x\rho x)$. \\\hline
				Symmetry&$\forall x,y\in X:\;x\rho y\Rightarrow y\rho x$ \\\hline
				Antisymmetry&$\forall x,y\in X:\;x\rho y \wedge y\rho x\Rightarrow x=y$ \\\hline
				Connecttivity&$\forall x,y\in X:\;x\rho y \vee y\rho x\vee x=y$ \\\hline
				Transitivity&$\forall x,y,z\in X:\;x\rho y \wedge y\rho z\Rightarrow x\rho z$
		\end{tabular}}
	\end{table}
	
	\subsection{Binary relation}\label{binary}
	For given sets $ X $ and $ Y $, the binary relation $ \rho $ defined over sets $ X $ and $ Y $ is any subset of the Cartesian product $ X\times Y $. If $ X=Y $, then the relation over the set $ X $ is a subset of the Cartesian product $ X\times X $ \cite{Gniazdowski2011}. Here, only relations over the n-element set $ X $ will be considered.
	
	A binary relation can have specific properties. Among others, a binary relation can be reflexive, irreflexive, symmetric, antisymmetric, connected, and transitive (Table \ref{tab1}). The properties of a binary relation determine its type. Several types of relations in particular are important from the point of view of data analysis:
	\begin{itemize}
		\item If a relation is reflexive, symmetric, and transitive, it is an equivalence relation. An equivalence relation divides a set into disjoint equivalence classes.
		\item If the relation is reflexive and symmetric, then it is a similarity relation. The similarity relation divides a set into similarity classes.
		\item If the relation is reflexive, antisymmetric and transitive, then it is a partial order relation. In a set with a partial order relation, ordering of a set (or sorting in the sense of a given relation) is possible only within certain subsets, and a set with a partial order relation is called a partially ordered set.
		\item If a relation is a partial order relation and at the same time is connected, then it is a linear order relation (or total order relation). A linear order is a stronger property than a partial order. In a set with a linear order relation, it is possible to fully ordering of a set.
	\end{itemize}
	In the set of real numbers $ R $, the following linear order relation is possible:
	\begin{equation}\label{eq1}
		\forall m,n\in R\;\; m\rho n\Leftrightarrow m\ge n.
	\end{equation}
	In the set of complex numbers $ C $, it is not possible to define the relation in the above way.
	Here it is possible to define the weaker relation, that is, a partial order relation:
	\begin{equation}\label{eq2}
		\forall m,n\in C\;\; m\rho n\Leftrightarrow (m=n)\vee (|m|>|n|).
	\end{equation}
	\subsection{Scales of measurement}
	Measurement (or data collection) assumes the existence of four measurement scales \cite{Stevens1946}:
	\begin{itemize}
		\item Nominal scale,
		\item Ordinal scale,
		\item Interval scale,
		\item Ratio scale. 
	\end{itemize}
	The nominal scale assumes the classification of data into different classes. For data measured on a nominal scale, it can be said that the two measured values are equivalent or different. It cannot be said that one value precedes another. This means that nominal data cannot be sorted in any way.
	
	Measurement on the ordinal scale is more precise than measurement on the nominal scale. The ordinal scale allows you to order a set according to the degree to which the elements of the set have certain features, but does not give information about the magnitude of the differences between these elements.
	
	The interval scale makes it possible not only to order objects in terms of the degree of possessing a certain feature, but also gives the ability to determine the distance between objects. An interval scale is a continuous numerical scale that has no absolute zero. Zero on the measurement scale is set arbitrarily. An example of interval scale measurement is a temperature measurement in degrees Celsius or degrees Fahrenheit. It is known how much one measurement result is greater than another. However, it is impossible to say how many times one result is greater than the other. For example, for temperature measurement on the Celsius scale, it can be said that the temperature of 27 degrees is 18 degrees higher than the temperature of 9 degrees. On the other hand, it cannot be said that a temperature of 27 degrees is three times higher than a temperature of 9 degrees.
	
	The ratio scale has all the features of an interval scale. Its additional feature is that this scale has absolute zero. Therefore, with regard to the measurement results on a ratio scale, it can be said how many times one measurement result is greater than another. An example of a ratio scale is the temperature measurement scale in degrees Kelvin.
	
	The concept of measurement scales is closely related to the concept of two types of relations. For elements of a set measured at a nominal scale, an equivalence relation can be defined. For elements of a set measured on an ordinal, interval and ratio scale, a linear order relation can be defined. Thanks to the linear order relation, data measured on ordinal, interval, and ratio scales can also be ranked.
	
	\begin{table}[t]
		\centering
		\caption{Sample contingency table}\label{tab2}
		\fontsize{9}{13}\selectfont{
			\begin{tabular}{ c||c|c||c }
				&$ X $&$ Y $&Total\\ \hline \hline
				$ A $&$ 3 $&$ 0 $&$ 3 $\\ \hline
				$B$&$2$&$2$&$4$\\ \hline
				$C$&$0$&$2$&$2$\\ \hline\hline
				Total&$5$&$4$&$9$\\ 
		\end{tabular}}
	\end{table}
	
	\begin{table}[t]
		\centering
		\caption{Random variables $ V_1 $ and $ V_2 $ identified from Table \ref{tab2}}\label{tab3}
		\fontsize{9}{13}\selectfont{
			\begin{tabular}{ c||c }
				$ V_1 $&$ V_2 $\\ \hline\hline
				A & X \\\hline
				A & X \\\hline
				A & X \\\hline
				B & X \\\hline
				B & X \\\hline
				B & Y \\\hline
				B & Y \\\hline
				C & Y \\\hline
				C & Y 
		\end{tabular}}
	\end{table}
	
	\subsection{The study of the relationship between nominal data in classical statistics}
	An example of a contingency table for nominal data (Table \ref{tab2}) is given. The table describes the interdependence of two nominal random variables  $ V_1 $ and $ V_2 $. The first variable ($ V_1 $) takes three values: $ \{A,B,C\} $. The second variable ($ V_2 $) takes two values: $ \{X,Y\} $. Table \ref{tab3} shows the random variables $ V_1 $ and $ V_2$ reconstructed from the contingency table (Table \ref{tab2}).
	
	\begin{table}[t]
		\centering
		\caption{Table of expected frequencies estimated from Table \ref{tab2}}\label{tab4}
		\fontsize{9}{13}\selectfont{
			\begin{tabular}{ c||c|c||c }
				&X&Y&Total\\  \hline \hline
				A&1.667&1.333&3\\ \hline
				B&2.222&1.778&4\\ \hline
				C&1.111&0.889&2\\ \hline \hline
				Total&5&4&9\\
		\end{tabular}}
	\end{table}
	
	\subsubsection{Statistics $ \chi^2 $}
	Individual cells of the contingency table ($ O_{i,j} $) count the observed frequencies of pairs of different nominal values. Based on the observed frequencies, the relevant elements ($ E_{ij} $) in the expected frequency table (Table \ref{tab4}) can be estimated \cite{Blalock1960}.  The $ E_{ij} $ element  is equal to the product of the sum of the elements in the i-th row of the contingency table $ (\sum_{j}O_{ij}) $ and the sum of the  elements in the j-th column of the contingency table $ (\sum_{i}O_{ij}) $, divided by  sum of all elements of the contingency table $ (\sum_{ij}O_{ij}) $:
	\begin{equation}
		E_{ij}=\frac{\sum_{j}O_{ij}\cdot\sum_{i}O_{ij}}{\sum_{ij}O_{ij}}.
	\end{equation}
	The estimated table of expected frequencies is shown in Table \ref{tab4}. Now, based on the contents of the table of observed frequencies and the table of expected frequencies, the value of the statistic $ \chi^2 $ can be estimated \cite{Blalock1960}:
	\begin{equation}
		\chi^2=\sum_{i.j}\frac{\left(O_{ij}-E_{ij}\right)^2}{E_{ij}}.
	\end{equation}
	For the contingency table considered here (Table \ref{tab2}), the statistic $ \chi^2 $ is $ 4.95 $. In classical statistics, the value of the $ \chi^2 $ statistic can be used to test the significance of the correlation between the two nominal variables. To do this, a null hypothesis should be formulated and then checked whether the null hypothesis can be rejected.
	\subsubsection{Testing the null hypothesis}
	The null hypothesis assumes that the examined variables are independent of each other, i.e. there is no correlation between the variables $ V_1 $ and $ V_2 $. The negation of the null hypothesis is the alternative hypothesis, which says that the variables $ V_1 $ and $ V_2 $ are significantly correlated. If the null hypothesis is rejected, the alternative hypothesis should be accepted instead.
	
	Denoting by r and c respectively the number of rows and the number of columns in the contingency table, the number of degrees of freedom of the test is obtained:
	\begin{equation}
		d_f=\left(r-1\right)\times\left(c-1\right).
	\end{equation}
	For the example considered here, $ d_f=\left(3-1\right)\left(2-1\right)=2 $. The probability $ p $ is estimated as the integral of the distribution function $ \chi^2 $:
	\begin{equation}
		p=\int_{\chi^2}^{\infty}f\left(x,d_f\right)dx.
	\end{equation}
	If the probability $ p $ is small enough, the null hypothesis can be rejected. Assuming that the significance level $ \alpha=0.1 $, the null hypothesis can be rejected when the obtained probability $ p $ is less than $ \alpha $. For the example in Table \ref{tab2}, the estimated probability $ p $ is $ 0.0841 $. This means that at the significance level $ \alpha=0.1 $, the null hypothesis should be rejected. At the same time, the alternative hypothesis, which says that both variables are significantly correlated, should be accepted.
	
	The $ \chi^2 $ test also has its limitations. When using this test, it is required that the expected frequencies be not less than five \cite{Blalock1960}. Some sources state that expected frequencies should not be less than ten \cite{Francuz2007}\cite{StatSoft2011}. If the expected numbers are too small, Yates' correction or Fisher's exact test \cite{Blalock1960}\cite{Francuz2007}\cite{StatSoft2011} is used for contingency tables of size $ 2\times2 $. For tables of larger sizes, Fisher's exact test is a combinatorial problem of high computational complexity. For such a case, a graph algorithm was proposed that extends the feasibility limits of Fisher's exact test \cite{Mehta1983}.
	\subsubsection{The strength of the correlation relationship}
	Regardless of the significance assessment, the $ \chi^2 $ statistic can also be used to estimate the strength of the correlation relationship. For this, you can use the V-Cramer coefficient, which takes values in the range $ [0,1] $. The square of the V-Cramer coefficient can be calculated using the formula:
	\begin{equation}\label{eq15}
		V^2=\frac{\chi^2}{N\cdot\min{\left(r-1,c-1\right)}}.
	\end{equation}
	In formula (\ref{eq15}), $ N $ is the number of observations. In the considered case $ N=9 $ and the calculated value of the coefficient $ V^2 $ is $ 0.550 $.
	
	\subsection{Numerical coding of nominal data}
	In statistics, in some cases, the numerical values of the observation results are replaced by their ranks. The ranking procedure consists in the fact that first the data set is sorted in non-decreasing order, and then its successive elements are assigned a rank equal to its position in the sorted set \cite{Wilcoxon1945}.
	
	It may happen that in a sorted set there will be elements with identical values. In this case, the ranks assigned to identical values should be the same. When a sorted set contains elements with identical values, these elements are assigned a rank equal to the average value of their positions in the sorted set \cite{Wilcoxon1945}. In this case, we talk about tied ranks.
	
	It should be emphasized that the value of assigned ranks depends on two factors. The first factor is the numerical value of the element in the set to be sorted. The second factor is the number of elements with equal values.
	
	In the case of nominal data, it is not possible to establish a linear order relation. A potential method of ranking nominal data could not use element sorting. On the other hand, intuition suggests that in a statistical collectivity, both in the case of numerical data and nominal data, an element with a higher cardinality is more important than an element with a lower cardinality. Therefore, a proposal arises to rank nominal variables, starting from the cardinality of their individual  values.
	\subsubsection{The case of classes with different cardinality}\label{rangi}
	A specific feature of nominal data is that it cannot be sorted by values, but can be clustered by identical values. In a given class containing $ n $ identical elements, these elements can be numbered with numbers from $ 1 $ to $ n $. By analogy with tied ranks, each of the identical elements can be assigned a rank equal to the average value of the assigned numbers:
	\begin{equation}\label{eq16}
		R=\frac{\left(n+1\right)}{2}.
	\end{equation}
	The rank of elements in the class with higher cardinality will be greater than the rank of elements belonging to the class of lower cardinality. Hence, it can be seen that in the set with ranks defined by the formula (\ref{eq16}) a linear order relation (\ref{eq1}) is possible.
	\subsubsection{The case of classes with equal cardinality}
	For classes with identical cardinalities, method (\ref{eq16}) would give identical ranks. This would lead to a situation where, after coding, elements belonging to classes with identical cardinalities would be indistinguishable. To prevent this, the method should be modified. If the nominal variable contains $ k $ different classes with identical cardinality, then the elements of the j-th class ($ j=0,1,\ldots,k-1 $) can be coded with $ k $ successive complex roots of unity:
	\begin{equation}
		R_j=R\cdot\sqrt[k]{1}=R\cdot e^{i\cdot\frac{2\pi j}{k}}=R\left(\cos{\frac{2\pi j}{k}}+i\sin{\frac{2\pi j}{k}}\right).
	\end{equation}
	In the above expression $ i=\sqrt{-1} $ , $ \varphi=2\pi j/k $   is the phase arbitrarily assigned to the successive (j-th) class, and $ R $ is the rank calculated according to the formula (\ref{eq16}) - the same as in the case of classes with different cardinalities. In the presented concept, $ R $ is a module of complex rank, depending on the cardinality of a given nominal value in the set. This approach assigns identical rank modules to classes with identical cardinalities, and distinguishes classes with identical cardinalities by their different phases.
	
	Unfortunately, unlike in a set with real ranks, a linear order relation is not possible in a set with complex ranks. Here only the partial order relation (\ref{eq2}) is possible.
	\subsubsection{Properties of numerical coding}
	Data coded with numbers (real or complex) acquire additional properties \cite{Gniazdowski2015}:
	\begin{itemize}
		\item Within a given variable, all classes are mutually distinguishable.
		\item The module of the number assigned to a given class contains information about the cardinality of the class, and thus about its statistical strength.
		\item In the phase of the complex number assigned to a given class, information about the number of classes with identical cardinality is contained.
	\end{itemize}
	Random variables gain the properties of vectors in numerical space (real or complex), on which arithmetic operations such as sum, difference, product, quotient, as well as exponentiation and root can be performed. In a complex vector space, it is also possible to define the scalar product:
	\begin{equation}
		\left(x,y\right)=\sum_{i=0}^{n}{x_i\overline{y_i}}.
	\end{equation}
	Based on the above definition of the  scalar product, the Euclidean norm can also be defined:
	\begin{equation}
		||x||=\sqrt{(x,x)}.
	\end{equation}
	If a norm is defined, then a metric can also be consistently defined:
	\begin{equation}
		\rho\left(x,y\right)=||y-x||.
	\end{equation}
	Since the real number is a special case of a complex number, the above formulas for the scalar product, the Euclidean norm and the metric are also valid when coding nominal variables with real numbers. And the ability to define a metric allows the use of numerical coding for the purposes of clustering and classification \cite{Gniazdowski2015}.
	
	\subsection{Geometric interpretation of the correlation coefficient}
	The measure of the relationship between two random variables is their covariance:
	\begin{equation}
		Cov(X,Y)=E[(X-\overline{X})(Y-\overline{Y})].
	\end{equation}
	The covariance normalized to unity is called the Pearson correlation coefficient:
	\begin{equation}
		R(X,Y)=\frac{Cov(X,Y)}{s_X s_Y}=\frac{E[(X-\overline{X})(Y-\overline{Y})]}{\sqrt{E[(X-\overline{X})^2]}\sqrt{E[(Y-\overline{Y})^2]}}.
	\end{equation}
	This expression can be further transformed to the form:
	\begin{equation}
		R(X,Y)=\frac{\sum_{i=1}^{n}[(X_i-\overline{X})(Y_i-\overline{Y})]}{\sqrt{\sum_{i=1}^{n}(X_i-\overline{X})^2}\sqrt{\sum_{i=1}^{n}(Y_i-\overline{Y})^2}}.
	\end{equation}
	Denoting the random components of both variables $ X $ and $ Y $ as $ x=X-\overline{X} $  and $ y=Y-\overline{Y} $, the formula for the correlation coefficient can be transformed into:
	\begin{equation}
		R(X,Y)=\frac{\sum_{i=1}^{n}{x_iy_i}}{\sqrt{\sum_{i=1}^{n}x_i^2}\sqrt{\sum_{i=1}^{n}y_i^2}}.
	\end{equation}
	In the numerator of the above formula there is the scalar product of the vectors $ x $  and $ y $, and in the denominator there is the product of the lengths of these vectors. This quotient expresses the cosine of the angle between the two vectors. This means that the correlation coefficient is identical to the cosine of the angle between the random components of random variables  \cite{Gniazdowski2013}:
	\begin{equation}\label{eq7}
		R(X,Y)=\frac{\sum_{i=1}^{n}{x_iy_i}}{\sqrt{\sum_{i=1}^{n}x_i^2}\sqrt{\sum_{i=1}^{n}y_i^2}}=\frac{x\cdot y}{||x||\cdot ||y||}=cos(\angle x,y).
	\end{equation}
	\subsubsection{Linear transformation of a random variable}\label{linear}
	A random variable $ X $ can be represented as the sum of its mean value $ \overline{X} $ and its random component $ x $:
	\begin{equation}
		X=\overline{X}+x.
	\end{equation}
	It can be seen that the linear transformation ($ a+bX $) of  a random variable $ X $ does not change the direction of the vector representing its random component:
	\begin{equation}
		a+bX=a+b(\overline{X}+x)=a+b\overline{X}+bx.
	\end{equation}
	The constant $ a+b\overline{X} $ represents the mean value of variable $ X $ after a linear transformation. The  random component $ x $ becomes  a new random component $ bx $ after the transformation of the variable $ X $. Since the vector $ bx $ is  parallel  to the vector $ x $, and the correlation coefficient is formally identical to the cosine of the angle between the vectors representing the random components of two random variables, this linear transformation does not change the modulus of the correlation coefficient.
	
	\subsection{Pearson correlation coefficient for complex random variables}
	By analogy, a method of calculating the correlation coefficient for complex random variables can be proposed. If $ x $ and  $ y $ are the random components of the  complex random variables $ X $ and $ Y $, then the correlation coefficient found as the cosine of the angle between the random components has the following form:
	\begin{equation}
		R=\frac{Cov(X,Y)}{\sqrt{var(X)var(Y)}}=\frac{E[x\overline{y}]}{\sqrt{E[x\overline{x}]E[y\overline{y}]}}=\frac{\sum_{i=1}^{n}x_i\overline{y_i}}{\sqrt{\sum_{i=1}^{n}x_i\overline{x_i}}\sqrt{\sum_{i=1}^{n}y_i\overline{y_i}}}.
	\end{equation}
	As noted in the introduction, in the formula above, $\overline{x}$ and $\overline{y}$ are conjugates of the complex numbers $x$ and $y$.
	\subsection{Least squares complex method}
	A linear model of $ \widehat{y} $ of a complex variable $ y $ with respect to complex variables $ x_1,x_2,\ldots,x_s $ is considered. The vector $ y=\left[y_1,y_2\ldots y_n\right]^T $ is the dependent variable of the model. The vector $ \widehat{y}=\left[\widehat{y}_1,\widehat{y}_2\ldots\widehat{y}_n\right]^T  $ is the estimate of the variable $ y $. For successive realizations of variables $ x_1,x_2,\ldots,x_s $ successive approximations of the vector $ y $ are obtained in the form of calculated values of $ \widehat{y} $:
	\begin{equation}
		\begin{matrix}\begin{matrix}{\widehat{y}}_1=b_0+b_1x_{11}+\ldots+b_sx_{1s}\\{\widehat{y}}_2=b_0+b_1x_{21}+\ldots+b_sx_{2s}\\\vdots\\\end{matrix}\\{\widehat{y}}_n=b_0+b_1x_{n1}+\ldots+b_sx_{ns}\\\end{matrix}.
	\end{equation}
	The above system of equations can be represented in matrix notation:
	\begin{equation}
		\widehat{y}=Xb.
	\end{equation}		
	In the above equation, the following notations were assumed:
	\begin{itemize}
		\item The matrix X represents the independent variables of the model:
		\begin{equation}
			X=\left[\begin{matrix}\begin{matrix}1&x_{11}&\ldots\\\vdots&\vdots&\vdots\\1&x_{n1}&\ldots\\\end{matrix}&\begin{matrix}x_{1s}\\\vdots\\x_{ns}\\\end{matrix}\\\end{matrix}\right].
		\end{equation}
		\item The vector $ b=\left[b_0,b_1\ldots b_s\right]^T $ is the vector of the model parameters.
		\item The vector $ e=\left[e_1,e_2\ldots e_n\right]^T=y-\widehat{y} $ is the error vector of the model.
	\end{itemize}
	The measure of model error is the quantity $ Q $, which is the square of the Euclidean norm of the error vector $ e $. Denoting the Hermitian transposition\footnote{The Hermitian transposition (or conjugate transposition) of a complex vector or complex matrix is understood as the composition of the transposition and conjugate of that vector or matrix, respectively.} of a vector (or matrix) $ M $ as $ M^H $, the measure of error of the model takes the form:
	\begin{equation}
		Q\left(b\right)=||e||^2=e^H\cdot e=(y-\widehat{y})^H\cdot (y-\widehat{y})=(y-Xb)^H\cdot(y-Xb).
	\end{equation}
	After multiplication, the error measure takes the form:
	\begin{equation}
		Q\left(b\right)=\left(y-Xb\right)^H\left(y-Xb\right)=y^Hy-y^HXb-b^HX^Hy+b^HX^HXb.
	\end{equation}
	It can be seen that the error measure Q (the square of the Euclidean norm) is the sum of the four real terms. On the other hand, $ y^HXb $ is a Hermitian transposition of $  b^HX^Hy $. Since the Hermitian transposition of a real number does not change its value, the equality holds:
	\begin{equation}
		y^HXb=b^HX^Hy.
	\end{equation}
	Therefore, the expression for $ Q\left(b\right) $ can be simplified a bit:
	\begin{equation}
		Q\left(b\right)=y^Hy-2b^HX^Hy+b^HX^HXb.
	\end{equation}
	The model error measure depending on model parameters $ b $ should be minimal. Hence:
	\begin{equation}
		\frac{\partial Q}{\partial b}=-2X^Hy+2X^HXb=0.
	\end{equation}
	After transformation, a system of normal equations is obtained \cite{Mok2014}\cite{whuber}:
	\begin{equation}\label{eq29}
		X^HXb=X^Hy.
	\end{equation}
	
	\begin{table}[t]
		\centering
		\caption{Nominal data from Table \ref{tab3} coded with numbers}\label{tab5}
		\fontsize{9}{13}\selectfont{
			\begin{tabular}{ c||c }
				$ V_1 $ & $ V_2 $ \\ \hline \hline
				2 & 3 \\ \hline
				2 & 3 \\ \hline
				2 & 3 \\ \hline
				2.5 & 3 \\ \hline
				2.5 & 3 \\ \hline
				2.5 & 2.5 \\ \hline
				2.5 & 2.5 \\ \hline
				1.5 & 2.5 \\ \hline
				1.5 & 2.5 \\
		\end{tabular}}
	\end{table}
	
	\section{Study of linear correlation between nominal random variables coded with numbers}\label{korelacje}
	Nominal data can be coded using real or complex numbers. On the one hand, the correlation coefficient between real random vectors has an interpretation of the cosine of the angle between their random components \cite{Gniazdowski2013}. On the other hand, for vectors in complex space, one can define the scalar product and the Euclidean norm, so one can also calculate the cosine of the angle between their random components. Similarly to the real space, this cosine can be interpreted as a correlation coefficient between complex random variables.
	\subsection{Example No. 1 - Nominal data coded with real numbers}\label{examp1}
	After coding the data from Table \ref{tab2}, their numerical representation was obtained as presented in Table \ref{tab5}. For the data from Table \ref{tab5}, the correlation coefficient $ R\left(V_1,V_2\right)=0.253 $ was obtained. 
	
	The problem was also investigated for many other datasets where both random variables could be coded with real numbers. Coding with real numbers introduced a linear order to the set of nominal data. In all analyzed cases, unambiguous real measures of linear correlation were obtained.
	
	\subsection{Nominal data coded with complex numbers}
	It can be seen that in Example No. 1 (Subsection \ref{examp1}), for a given variable, its different classes had different cardinalities. That is, they could be coded with different real numbers. It should be investigated whether the existence of classes with equal cardinality also allows the estimation of correlations between variables.
	
	\begin{table}[t]
		\centering
		\caption{Contingency table for nominal data with a random variable $ V_1 $ containing two classes ($ A $ and $ B $) with identical cardinalities}\label{tab6}
		\fontsize{9}{13}\selectfont{
			\begin{tabular}{ c||c|c||c }
				&X&Y&Total\\ \hline \hline
				A&3&1&4\\ \hline
				B&2&2&4\\ \hline \hline
				Total&5&3&8\\ 
		\end{tabular}}
	\end{table}
	
	\begin{table}[t]
		\centering
		\caption{Observation table identified from Table \ref{tab6}}\label{tab7}
		\fontsize{9}{13}\selectfont{
			\begin{tabular}{ c||c }
				$ V_1 $&$ V_2 $\\ \hline \hline
				A&X \\ \hline
				A&X \\ \hline
				A&X \\ \hline
				A&Y \\ \hline
				B&X \\ \hline
				B&X \\ \hline
				B&Y \\ \hline
				B&Y \\
		\end{tabular}}
	\end{table}
	
	\begin{table}[t]
		\centering
		\caption{Data from Table \ref{tab7} after complex coding of variables $ V_1 $ and $ V_2 $. In the columns $ V_{11} $ and $ V_{12} $ two phase permutations are taken into account }\label{tab8}
		\fontsize{9}{13}\selectfont{
			\begin{tabular}{ c|c||c }
				$ V_{11} $&$ V_{12} $&$ V_2 $\\ \hline \hline
				$ -2.5+0i $&$ 2.5+0i $& 3 \\ \hline
				$ -2.5+0i $&$ 2.5+0i $&3\\ \hline
				$ 	-2.5+0i $&$ 	2.5+0i $&3\\ \hline
				$ -2.5+0i $&$ 2.5+0i $&2\\ \hline
				$ 2.5+0i $&$ -2.5+0i $&3\\ \hline
				$ 2.5+0i $&$ -2.5+0i $&3\\ \hline
				$ 2.5+0i $&$ -2.5+0i $&2\\ \hline
				$ 2.5+0i $&$ -2.5+0i $&2\\
		\end{tabular}}
	\end{table}
	
	\begin{figure}[t]
		\centering
		\includegraphics[width=0.65\textwidth]{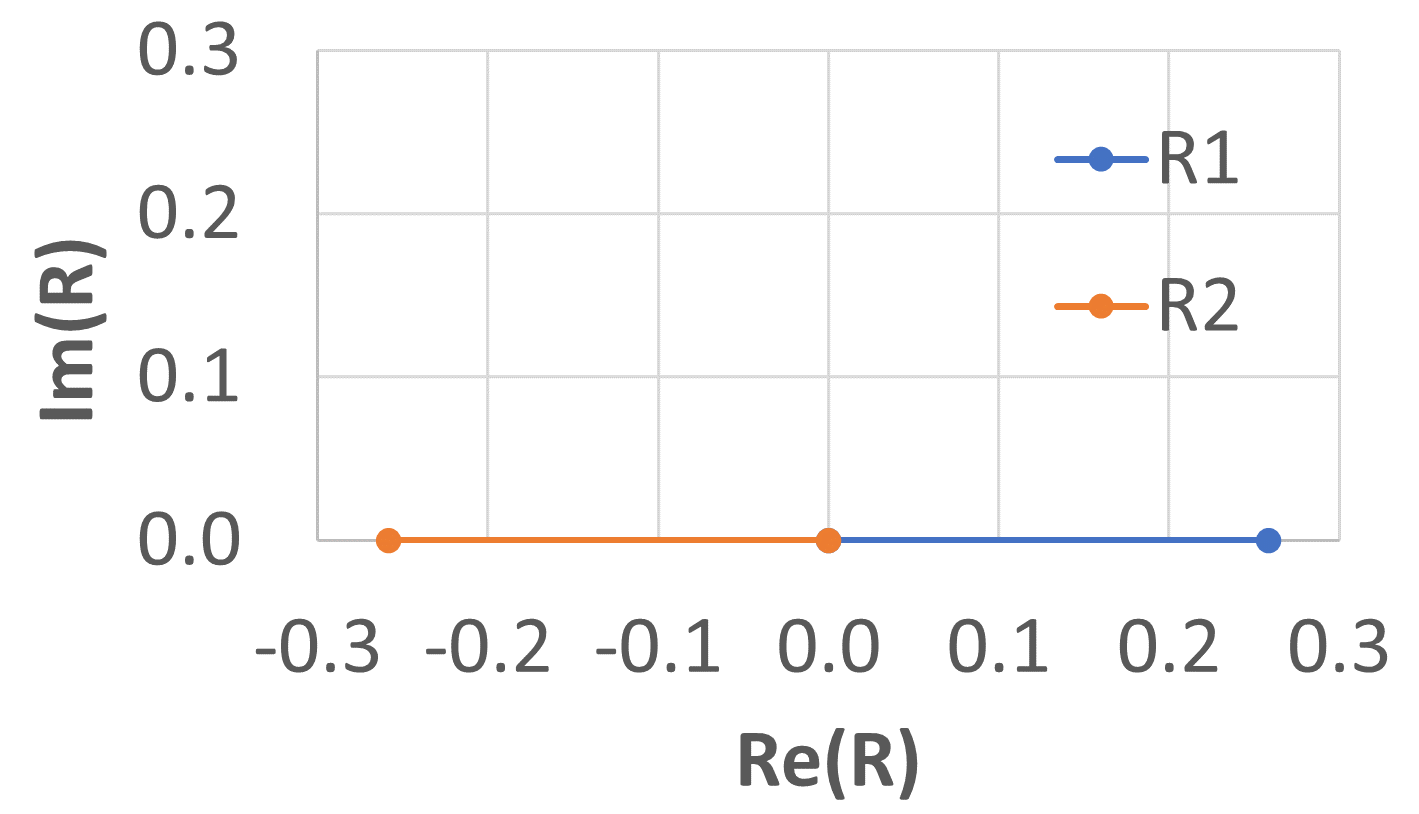}
		\caption{The correlation coefficients between variables $V_1$ and $V_2$ presented in the complex plane for the coded data from Table \ref{tab7}, obtained for $2!=2$ different phase permutations}\label{fig1}
	\end{figure}
	
	\subsubsection{Example No. 2 – nominal variable with two classes with identical cardinalities}\label{examp2}
	Table \ref{tab6} shows an example of a contingency table in which the first nominal variable contains two classes with identical cardinalities. These are classes containing values $ A $ and $ B $. Table \ref{tab7} shows the reconstructed random variables $ V_1 $ and $ V_2 $. Since the variable $ V_1 $ takes the value of $ A $ four times and also takes the value of $ B $ four times, the modulus of numbers used to code these values is $ 2.5 $. On the other hand, the arbitrarily adopted phase allowing to distinguish the values of $ A $ and $ B $ after their complex coding is $ \varphi=0^0 $ and $ \varphi={180}^0 $, respectively. 
	
	Since the phase for complex coding is chosen arbitrarily, the question arises about the influence of the selected phase on the value of the correlation coefficient. For two classes with identical cardinalities, the phases can be chosen in $ 2!=2 $ ways. Both ways of coding a random variable $ V_1 $ are shown in the first two columns of Table \ref{tab8} ($ V_{11} $ and $ V_{12} $).  The third column shows the variable $ V_2 $ after coding. For the data from Table \ref{tab8}, correlation coefficients have been estimated, which are $ R\left(V_1,V_2\right)=-0.258 $ and $ R\left(V_1,V_2\right)=0.258 $, respectively.
	
	The correlation coefficients for different phase permutations are shown in Figure \ref{fig1}. Comparing both values of the estimated correlation coefficients, it should be noted that their values result from the arbitrarily assigned phase in the coding of the nominal variable with two classes with identical cardinalities. It can be seen that both values of the correlation coefficient are located on the real axis symmetrically with respect to the center of the coordinate system. This means that the obtained solution is not unambiguous.
	
	\begin{table}[t]
		\centering
		\caption{Contingency table for nominal data with a random variable containing three classes ($ A $, $ B $ and $ C $) with identical cardinalities}\label{tab9}
		\fontsize{9}{13}\selectfont{
			\begin{tabular}{ c||c|c||c }
				&$ X $&$ Y $&Total\\ \hline \hline
				A&1&4&5\\ \hline
				B&2&3&5\\ \hline
				C&3&2&5\\ \hline
				D&1&2&3\\ \hline \hline
				Total&7&11&18\\  
		\end{tabular}}
	\end{table}
	
	\begin{table}[t]
		\centering
		\caption{Data from the Table \ref{tab9} after coding variables $V_1$ and $ V_2 $. The first six columns show the representations of the $ V_1 $ variable obtained for different phase permutations}\label{tab10}
		\fontsize{9}{13}\selectfont{
			\begin{tabular}{ c|c|c|c|c|c||c }
				$V_{11}$&$V_{12}$&$V_{13}$&$V_{14}$&$V_{15}$&$V_{16}$&$V_2$\\ \hline\hline
				3&-1.5+2.6i&-1.5-2.6i&3&-1.5+2.6i&-1.5-2.6i&4\\\hline
				3&-1.5+2.6i&-1.5-2.6i&3&-1.5+2.6i&-1.5-2.6i&6\\\hline
				3&-1.5+2.6i&-1.5-2.6i&3&-1.5+2.6i&-1.5-2.6i&6\\\hline
				3&-1.5+2.6i&-1.5-2.6i&3&-1.5+2.6i&-1.5-2.6i&6\\\hline
				3&-1.5+2.6i&-1.5-2.6i&3&-1.5+2.6i&-1.5-2.6i&6\\\hline
				-1.5+2.6i&3&3&-1.5-2.6i&-1.5-2.6i&-1.5+2.6i&4\\\hline
				-1.5+2.6i&3&3&-1.5-2.6i&-1.5-2.6i&-1.5+2.6i&4\\\hline
				-1.5+2.6i&3&3&-1.5-2.6i&-1.5-2.6i&-1.5+2.6i&6\\\hline
				-1.5+2.6i&3&3&-1.5-2.6i&-1.5-2.6i&-1.5+2.6i&6\\\hline
				-1.5+2.6i&3&3&-1.5-2.6i&-1.5-2.6i&-1.5+2.6i&6\\\hline
				-1.5-2.6i&-1.5-2.6i&-1.5+2.6i&-1.5+2.6i&3&3&4\\\hline
				-1.5-2.6i&-1.5-2.6i&-1.5+2.6i&-1.5+2.6i&3&3&4\\\hline
				-1.5-2.6i&-1.5-2.6i&-1.5+2.6i&-1.5+2.6i&3&3&4\\\hline
				-1.5-2.6i&-1.5-2.6i&-1.5+2.6i&-1.5+2.6i&3&3&6\\\hline
				-1.5-2.6i&-1.5-2.6i&-1.5+2.6i&-1.5+2.6i&3&3&6\\\hline
				2&2&2&2&2&2&4\\\hline
				2&2&2&2&2&2&6\\\hline
				2&2&2&2&2&2&6\\
		\end{tabular}}
	\end{table}
	
	\begin{figure}[t]
		\centering
		\includegraphics[width=0.80\textwidth]{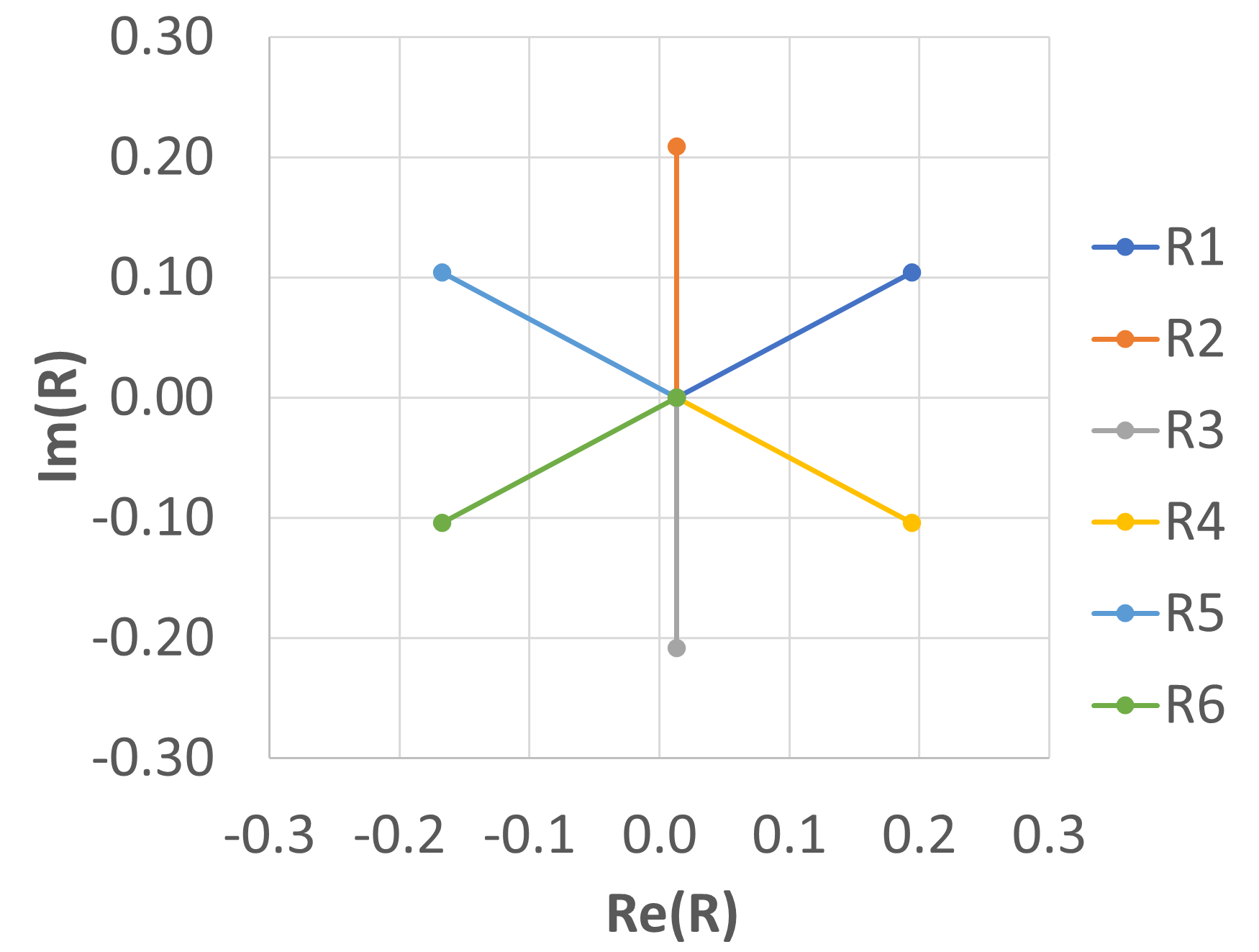}
		\caption{The correlation coefficients between variables $V_1$ and $V_2$ presented in the complex plane for the coded data from Table \ref{tab10}, obtained for $3!=6$ different phase permutations}\label{fig2}
	\end{figure}
	
	\subsubsection{Example No. 3 - nominal variable with three classes with identical cardinalities}\label{examp3}
	Another example of data is presented in Table \ref{tab9}. Variable $ V_1 $ contains three classes of elements with equal cardinality. These are classes $ A $, $ B $ and $ C $. There is also a class $ D $ with cardinality other than classes $ A $, $ B $ and $ C $. Classes $ A $, $ B $ and $ C $ contain $ 5 $ elements each. Class $ D $ contains $ 3 $ elements. The nominal value of $ D $ was coded with the real number $ 2=\left(3+1\right)/2 $. The nominal values $ A $, $ B $ and $ C $ are assigned the modulus $ 3=(5+1)/2 $. An arbitrary phase allowing to distinguish the coded values of $ A $, $ B $ and $ C $  can take one of three values $ \varphi=0^0 $, $ \varphi={120}^0 $ and $ \varphi={240}^0 $, respectively. In table \ref{tab10}, the first six columns show the coded prime variable for all $ 6=3! $ phase permutations. The second index digit in the variable name $ V_1 $ is used to distinguish permutations. The last column of Table \ref{tab10} also shows the variable $ V_2 $ after coding.
	
	For the data in Table \ref{tab10}, six complex correlation coefficients between the $ V_1 $ and $ V_2 $ variables were estimated, for different phase permutations attributed to different values of the $ V_1 $ variable belonging to classes having identical cardinalities. The obtained results are presented in the complex plane in Figure \ref{fig2}. It can be seen that the obtained correlation coefficients are distributed in the complex plane symmetrically with respect to the point with the coordinate $ 0.013 $ lying on the real axis. In contrast to the previous example, the center of symmetry for all obtained correlation coefficients is outside the central point of the complex plane. This is because in the current example, the variable $ V_1 $, in addition to the three equal-cardinality classes containing the values $ A $, $ B $, and $ C $, also has a class $ D $ with a cardinality different from classes $ A $, $ B $, and $ C $. As in the previous example, the solution obtained is not unambiguous.
	
	\begin{table}[t]
		\centering
		\caption{Contingency table for data with random variable $V_1$ containing four classes ($ U $, $ W $, $ X $ and $ Y $) with the same cardinality}\label{tab11}
		\fontsize{9}{13}\selectfont{
			\begin{tabular}{ c||c|c|c|c||c }
				&A&B&C&D&Total\\ \hline \hline
				U&45&20&13&12&90\\ \hline
				W&15&45&20&10&90\\ \hline
				X&5&8&50&27&90\\ \hline
				Y&8&10&17&55&90\\ \hline
				Z&10&30&50&450&540\\ \hline \hline
				Total&83&113&150&554&900
		\end{tabular}}
	\end{table}
	
	\begin{figure}[t]
		\centering
		\includegraphics[width=0.99\textwidth]{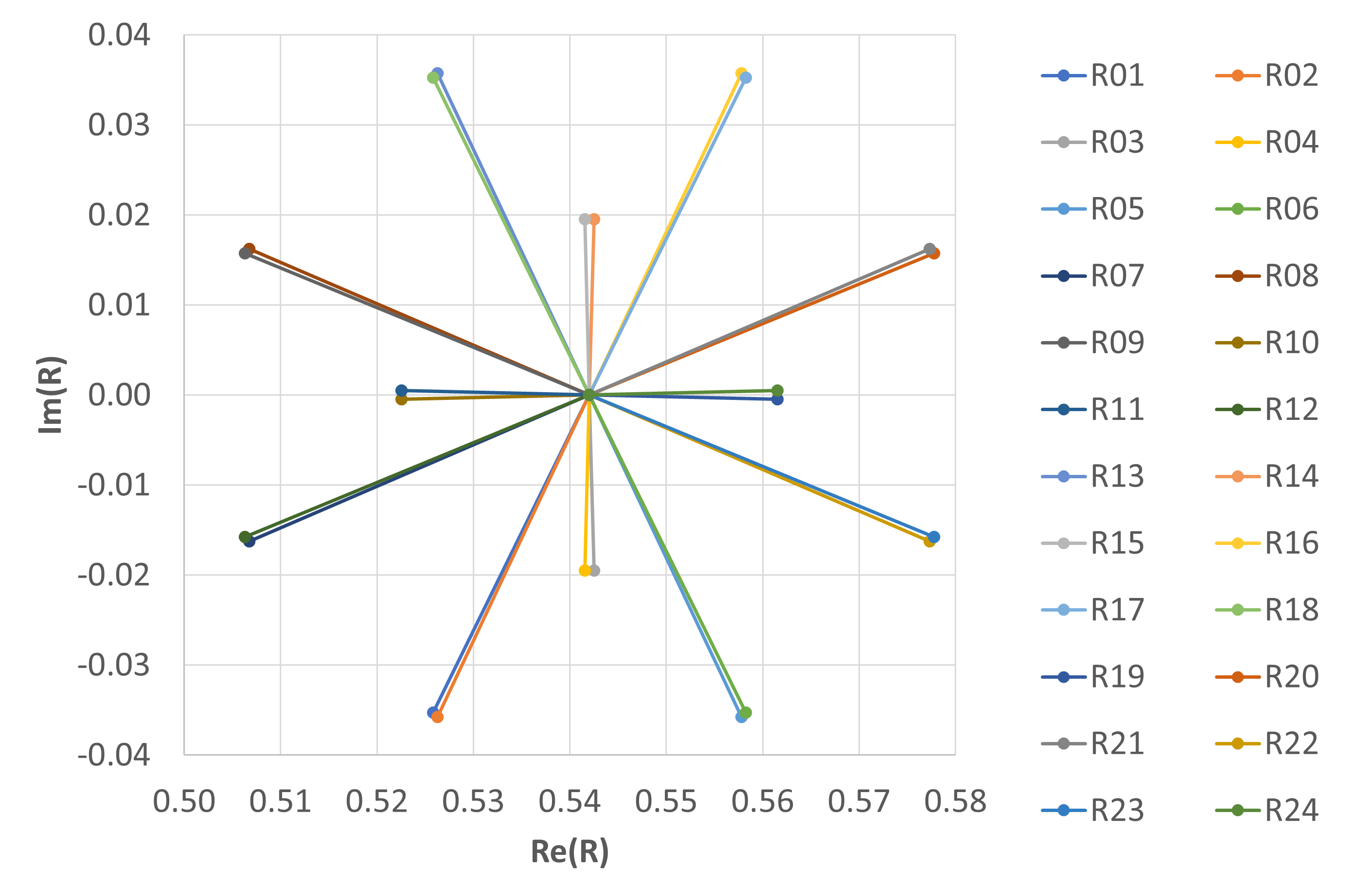}
		\caption{The correlation coefficients between variables $V_1$ and $V_2$ presented in the complex plane for the coded data from Table \ref{tab11}, obtained for $ 4!=24 $ different phase permutations}\label{fig3}
	\end{figure}
	
	\subsubsection{Example No. 4 - nominal variable with four classes with identical cardinalities}\label{examp4}
	Another example of data is shown in Table \ref{tab11}. The variable $ V_1 $ contains four classes with equal cardinality, containing the values of $ U $, $ W $, $ X $ and $ Y $. This variable also contains a class of $ Z $ values whose cardinality is different from the cardinality of classes $ U $, $ W $, $ X $ and $ Y $. The variable $ V_1 $ was coded using complex numbers. Since the variable $ V_1 $ contains four classes with the same cardinality, $ 4!=24 $ different phase permutations are possible. For all these permutations, Figure \ref{fig3} shows all possible values of correlation coefficients between variable $ V_1 $ and variable $ V_2 $ in the complex plane. Since the variable $ V_1 $ contains a class of $ Z $ values with a cardinality different from the cardinality of classes $ U $, $ W $, $ X $ and $ Y $, therefore also this time the estimated correlation coefficients are not symmetric with respect to the center of the coordinate system. The results are distributed symmetrically around a point with the coordinate $ 0.542 $ on the real axis. In this case, too, no unambiguous solution was obtained.
	
	\begin{table}[t]
		\centering
		\caption{Contingency table for data with random variable $V_1$ containing five classes ($ A $, $ B $, $ C $, $ D $ and $ E $) with the same cardinality}\label{tab12}
		\fontsize{9}{13}\selectfont{
			\begin{tabular}{ c||c|c|c||c }
				&X&Y&Z&Total\\ \hline \hline
				A&200&120&130&450\\ \hline
				B&250&100&100&450\\ \hline
				C&50&80&320&450\\ \hline
				D&170&130&150&450\\ \hline
				E&300&100&50&450\\ \hline \hline
				Total&970&530&750&2250
		\end{tabular}}
	\end{table}
	
	\begin{figure}[t]
		\centering
		\includegraphics[width=0.99\textwidth]{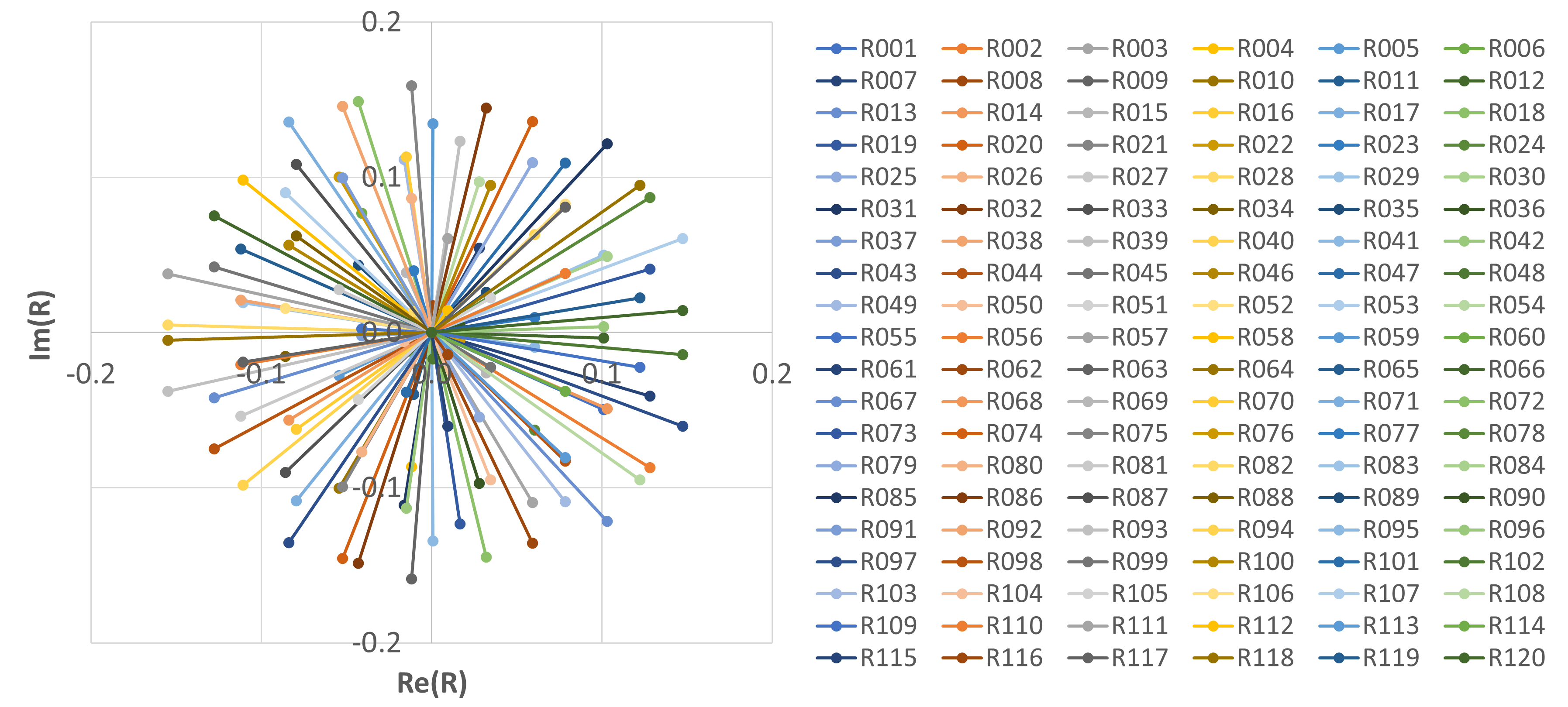}
		\caption{The correlation coefficients between variables $V_1$ and $V_2$ presented in the complex plane for the coded data from Table \ref{tab12}, obtained for $ 5!=120 $ different phase permutations}\label{fig4}
	\end{figure}
	
	\subsubsection{Example No. 5 – nominal variable with five classes with identical cardinalities}\label{examp5}
	Another example of nominal data is shown in Table \ref{tab12}.  In the considered example, the first variable takes $ 450 $ times each of the values from the set $ \{A,B.C,D,E\} $. For five classes with equal cardinality, the number of different codings resulting from the phase permutations is $ 5!=120 $. For all these cases, complex correlation coefficients were estimated between both variables.  $ 120 $ different correlation coefficients were obtained, which are presented in the complex plane in Figure \ref{fig4}. Since all classes contained in the set of values of the first variable have equal cardinality, therefore the obtained correlation coefficients are distributed symmetrically with respect to the center of the coordinate system.
	\section{Linear correlation ambiguity problem with complex coding of nominal data}
	The examples show that for nominal data coded with complex numbers, the problem of finding the linear correlation coefficient is ambiguous.
	Complex coding of $ k $ equal-cardinality classes assumes that each of these classes is coded with a complex number of equal modulus and with an arbitrarily chosen phase equal to the phase of one of the $ k $ roots of unity.    Therefore, $ k $ classes of equal cardinality can be assigned phases in $ k! $ ways. This means that a nominal random variable can be represented in $ k! $ ways. For these $ k $! possible representations of the first random variable and a unique representation of the second random variable, $ k! $ different complex correlation coefficients are obtained. In the complex plane, these correlation coefficients are symmetrical about some point on the real axis.
	
	In real space, the correlation coefficient is identical to the cosine of the angle between the random components of two random variables. Calculating the cosine of an angle in a complex space and treating it as a correlation coefficient, due to ambiguity, does not lead to a satisfactory result.
	\subsection{Searching for invariants with respect to phase permutations in complex coding of nominal data}
	Since the measure of linear correlation cannot be the cosine of the angle between the vectors representing the random components of complex random variables, an attempt was also made to find alternative correlation measures that would be invariant with respect to phase permutations between classes with equal cardinalities. 
	For this purpose, a linear correlation between the $ V_2 $ variable and its model $ \widehat{V_2}=f\left(V_1\right) $ was considered. First, tests were carried out for the linear model $ f(V_1) $, and then for the non-linear model.
	
	\begin{table}[t]
		\centering
		\caption{ Comparison between the correlation coefficients $R\left(V_1,V_2\right)$ and $R\left(V_2,\widehat{V_2}\right)$ for the coded data from Table \ref{tab10}, obtained for $3!=6$ different phase permutations}\label{tab13}
		\fontsize{9}{13}\selectfont{
			\begin{tabular}{ c|c|c|c|c }
				No.&$ R\left(V_1,V_2\right) $&$ |R\left(V_1,V_2\right)|$&$ \widehat{V_2}=b_0+b_1  V_1 $&$ R\left(V_2,\widehat{V_2}\right) $ \\ \hline \hline
				1&0.194+0.104i&0.220&$\left(5.200+0.012i\right)+\left(0.067+0.036i\right)  V_1$&0.220\\ \hline
				2&0.013+0.209i&0.209&$\left(5.221+0.024i\right)+\left(0.005-0.072i\right)  V_1$&0.209\\ \hline
				3&0.013-0.209i&0.209&$\left(5.221-0.024i\right)+\left(0.005+0.072i\right)  V_1$&0.209\\ \hline
				4&0.194-0.104i&0.220&$\left(5.200-0.012i\right)+\left(0.067+0.036i\right)  V_1$&0.220\\ \hline
				5&-0.167+0.104i&0.197&$\left(5.241+0.012i\right)-\left(0.058+0.036i\right)  V_1$&0.197\\ \hline
				6&-0.167-0.104i&0.197&$\left(5.241-0.012i\right)-\left(0.058-0.036i\right)  V_1$&0.197
		\end{tabular}}
	\end{table}
	
	\subsubsection{Correlation between variable $ V_2 $ and linear model $ \widehat{V_2}=f\left(V_1\right) $}
	In the considered cases, the data were selected in such a way that the first (only the first) random variable had classes with identical cardinality. For these classes, all phase permutations were taken into account in their complex coding. For many pairs of random variables, the least squares method was used to identify linear models of variable $ V_2 $ with respect to the complex variable $ V_1 $. To solve the complex system of normal equations (\ref{eq29}, the Gaussian elimination algorithm with partial pivoting was used \cite{Dryja1982}\cite{Kielbasinski1992}. In all cases, complex correlations between the $ V_2 $ variable and its $ \widehat{V_2} $ model were estimated. These correlations depended on phase permutations, and the correlation modulus was identical to the linear correlation modulus between variables $ V_1 $ and $ V_2 $.
	
	Table \ref{tab13} shows an example of the data from Table \ref{tab9}. 
	The table compares the correlation coefficients$  R\left(V_1,V_2\right) $ obtained for all phase permutations from Table \ref{tab10} with the correlation coefficients $ R\left(V_2,\widehat{V_2}\right) $ obtained for the same phase permutations.
	
	As a result of the research, it was found that the correlation between the variable and its linear model is not invariant with respect to phase permutations. Therefore, it cannot be considered as a practical measure of the correlation between a nominal random variable and a random variable measured on a nominal scale or on a stronger measurement scale.
	
	\begin{figure}[t]
		\centering
		{\subfloat[] {\label{figure5a}
				\includegraphics[width=0.475\textwidth]{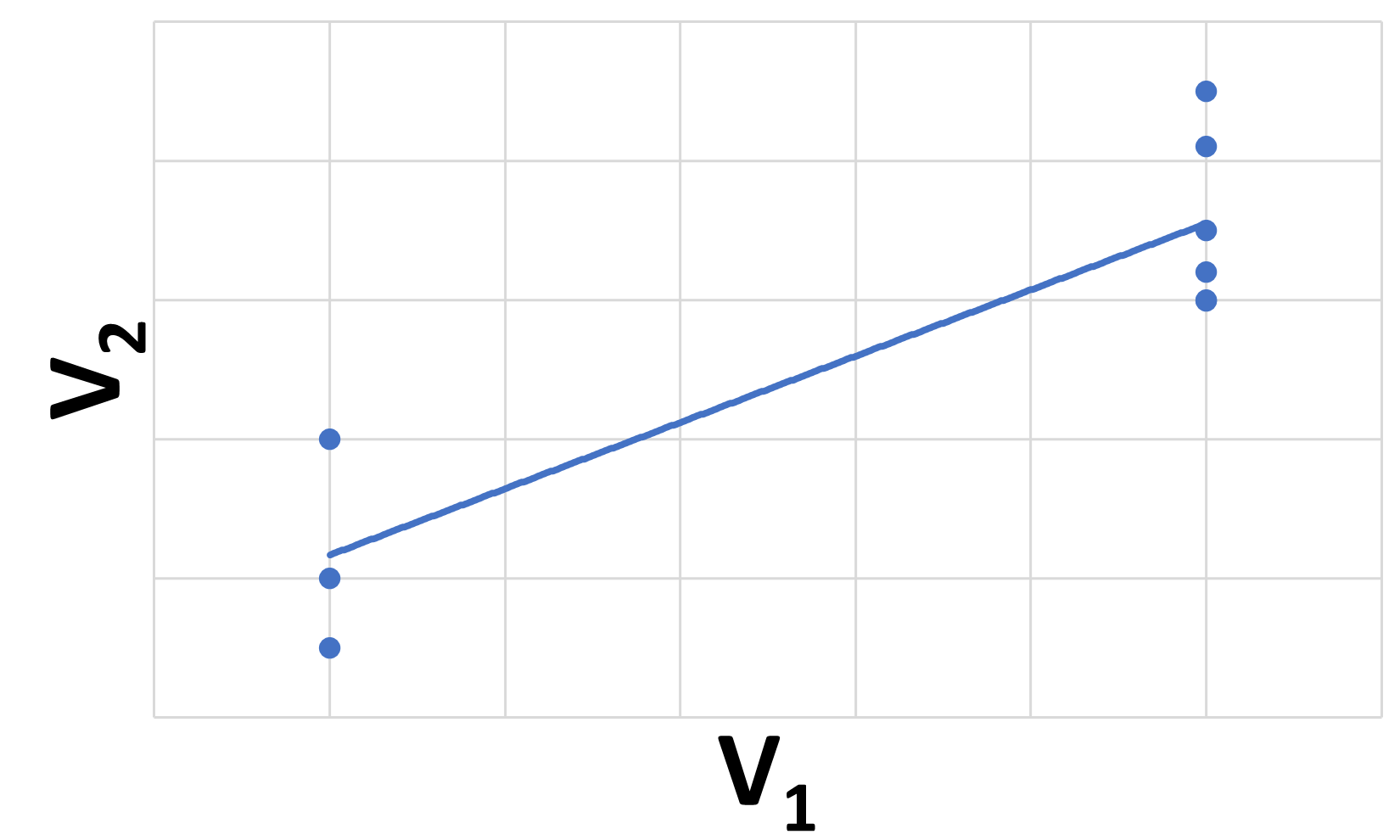}
			} \,
			\subfloat[] {\label{figure5b}
				\includegraphics[width=0.475\textwidth]{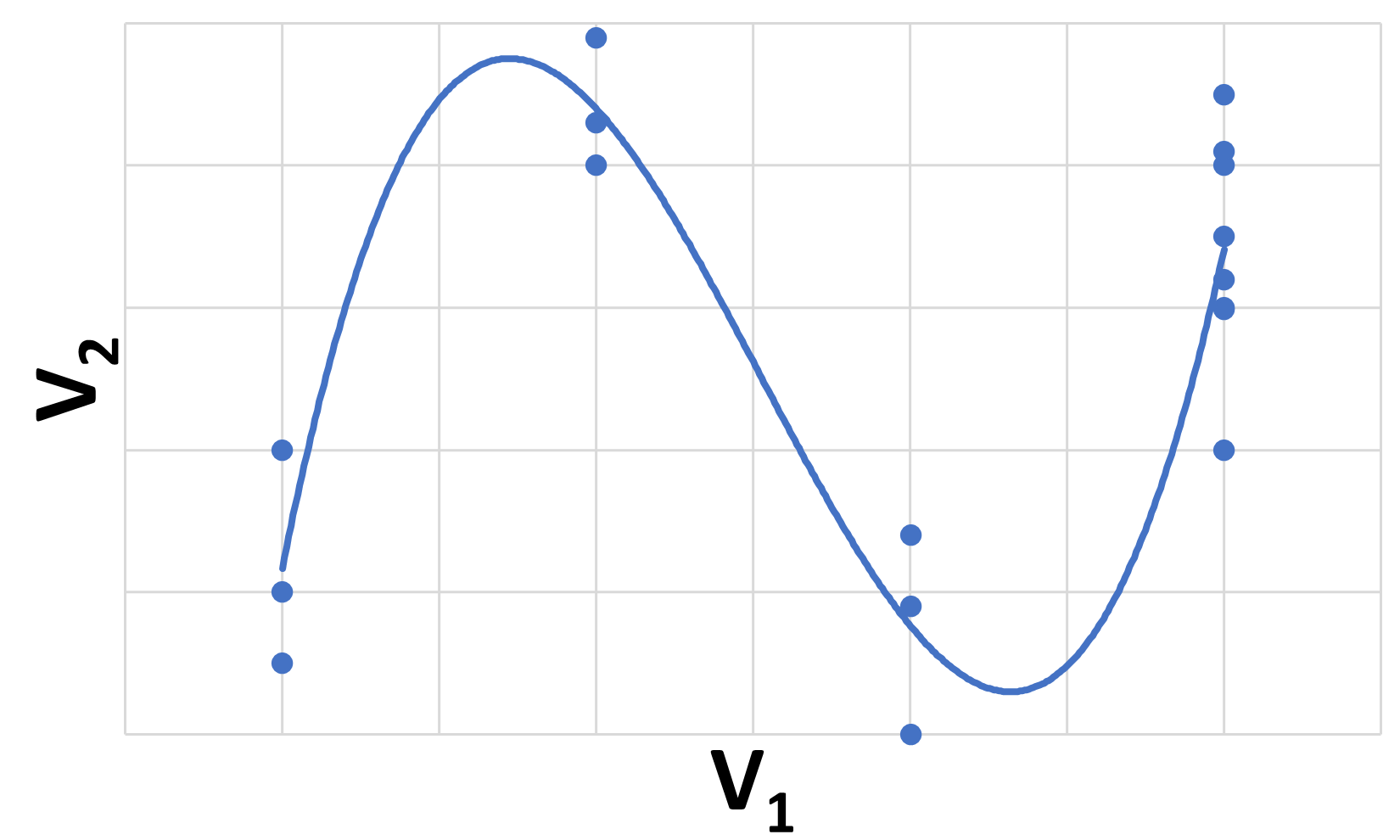}
			}
			\caption{Approximation of the relationship $ \widehat{V_2}=f\left(V_1\right) $ for $ k $ of different values contained in the variable $ V_1 $: (a) straight line for $ k=2 $; (b) a polynomial of the third degree for $  k=4 $}\label{fig5}}
	\end{figure}
	
	\begin{table}[t]
		\centering
		\caption{Comparison of correlation coefficients between variable $V_2$ and its non-linear model $\widehat{V_2}=f(V_1)$ for the coded data from Table \ref{tab10}, obtained for $3!=6$ different phase permutations}\label{tab14}
		\fontsize{8}{13}\selectfont{
			\begin{tabular}{ c|c|c }
				No.&$\widehat{V_2}=b_0+b_1V_1+b_2V_1^2+b_3V_1^3$&$R\left(V_2,\widehat{V_2}\right)$\\ \hline \hline
				1&$\left(5.074+0.036i\right)+\left(0.067-0.038i\right)  V_1+\left(0.022+0.013i\right)  V_1^2+\left(0.005-0.001i\right)  V_1^3$&0.31\\ \hline
				2&$\left(5.389+0.073i\right)+\left(0.000-0.077i\right)  V_1+\left(0.000+0.026i\right)  V_1^2-\left(0.007+0.003i\right)  V_1^3$&0.31\\ \hline
				3&$\left(5.389-0.073i\right)+\left(0.000+0.077i\right)  V_1+\left(0.000-0.026i\right)  V_1^2-\left(0.007-0.003i\right)  V_1^3$&0.31\\ \hline
				4&$\left(5.074-0.036i\right)+\left(0.067+0.038i\right)  V_1+\left(0.022-0.013i\right)  V_1^2+\left(0.005+0.001i\right)  V_1^3$&0.31\\ \hline
				5&$\left(5.705+0.036i\right)-\left(0.067+0.038i\right)  V_1-\left(0.022-0.013i\right)  V_1^2-\left(0.019+0.001i\right)  V_1^3$&0.31\\ \hline
				6&$\left(5.705-0.036i\right)-\left(0.067-0.038i\right)  V_1-\left(0.022+0.013i\right)  V_1^2-\left(0.019-0.001i\right)  V_1^3$&0.31
		\end{tabular}}
	\end{table}
	
	\subsubsection{Correlation between variable $ V_2 $ and non-linear model $ \widehat{V_2}=f\left(V_1\right)$}\label{nonlin}
	Due to the  resulting ambiguities of  linear correlation, another attempt was made to find a measure used to assess the strength of the correlation relationship. This time the non-linear correlation was studied.
	
	If the nominal variable $ V_1 $ contains $ k $ classes, then after numerical coding this variable takes $ k $ of different complex or real values. Each value of the variable $ V_1 $ corresponds to some (real) values representing the variable $ V_2 $. The relationship $ \widehat{V_2}=f\left(V_1\right) $ can be described by a polynomial of degree $ k-1 $. This relationship can be identified by the method of least squares. This is symbolically illustrated by the examples in Figure \ref{fig5}. For $ k=2 $, the relationship is approximated by a straight line (Figure \ref{fig5}(a)), and for $ k=4 $ it is a polynomial of the third degree (Figure \ref{fig5}(b)).
	
	By finding the correlation coefficient between the values calculated using the $ \widehat{V_2}=f\left(V_1\right) $ model and the numerical values representing the $ V_2 $ variable, the non-linear correlation between the $ V_1 $ variable and the $ V_2 $ variable can be estimated. If among all $ k $ different values of the variable $ V_1 $ there are $ m $ classes with equal cardinality ($ m\le k $), then these classes are coded with complex numbers. For all these classes, the moduli of the complex numbers are identical. To distinguish between classes of equal cardinality, modules were arbitrarily assigned phases of successive roots of degree $ m $ of unity. For different phase permutations, different coefficients in the identified polynomial will also be obtained. The question remains whether the correlation coefficients between the values calculated using the $ \widehat{V_2} $ model and the numerical values representing the $ V_2 $ variable change for different phase permutations.
	
	The problem posed in this way was tested for many different data sets in which there were classes with equal cardinality. Variables with two, with three, with four and also with five classes with the same cardinalities were tested. For each data set, for all phase permutations, non-linear models $ \widehat{V_2}=f(V_1) $ were identified using the least squares method.  For two, three, four and five identical cardinalities, $ 2!=2 $, $ 3!=6 $, $ 4!=24 $ and $ 5!=120 $ polynomials were identified, respectively. 
	In all tested cases, regardless of the phase permutations for the coded nominal values of the random variable, the linear correlation coefficient $ R(V_2,\widehat{V_2}) $ between the non-linear polynomial $ \widehat{V_2}=f(V_1) $, and the variable $ V_2 $ did not change with successive permutations. This means that the measure obtained here is invariant with respect to phase permutations in the complex coding of a random variable. Unfortunately, the found measure measures the non-linear correlation between the variables $ V_1 $ and $ V_2 $. For this reason, in those cases where a linear measure of correlation is expected, the non-linear correlation coefficient above could not be used.
	
	Table \ref{tab14} shows examples of results obtained for the data in Table \ref{tab9}.    In the dataset, variable $ V_1 $ contained four different classes, three of which had equal cardinality. Hence, the variable $ V_2 $ could be modeled using a third degree polynomial. For three classes with equal cardinality, $ 3!=6 $ different polynomials were obtained.	
	\section{Conclusions}
	The aim of the work was to examine the possibility of measuring the strength of the linear correlation relationship between two random variables measured on a nominal scale, coded with real numbers or complex numbers. The research was conducted with the assumption that the second random variable will always be coded with real numbers. This assumption caused the analyzed problem to become equivalent to the problem of testing the strength of a linear correlation relationship between a random variable measured on a nominal scale and a random variable measured at least on an ordinal scale. The calculations made use of the fact that the correlation coefficient has an interpretation of the cosine of the angle between the vectors containing the random components of the analyzed variables. Since for vectors of numbers  (both real and complex) Euclidean norms and the scalar product can be calculated, it is also possible to calculate the cosine of the angle between vectors, and consequently also to estimate the correlation coefficient between them. The conducted research allowed to draw several conclusions that may potentially be useful in the analysis of linear correlation for nominal data, and which will be presented in more detail in the following subsections.
	\subsection{Study of the correlation relationship between nominal data coded with real numbers}
	A set of data measured on a nominal scale may be coded using real numbers if there are no classes with identical cardinalities among the classes of identical elements contained in this set. In this case, correlations were studied for two variables, each of which could be coded with real numbers. Coding with real numbers introduced a linear order to the nominal data sets. As a consequence, unambiguous real measures of linear correlation were obtained in all analyzed cases.
	\subsection{Study of the correlation relationship between nominal data coded with complex numbers}
	According to the adopted assumptions, correlations between two random variables were investigated, one of which was coded with real numbers, and the other had to be coded with complex numbers. A set of data measured on a nominal scale cannot be coded with real numbers if there are at least two classes with identical cardinalities among the classes of identical elements contained in this set. For two random variables, one of which contained complex numbers and the other contained real numbers, complex correlation coefficients were obtained, which changed with the permutation of phases in complex numbers coding classes of elements with equal cardinality. The solution to the problem of finding linear correlation turned out to be ambiguous.
	\subsection{Necessary condition for linear correlation}
	The correlation coefficient can only be estimated for real random variables, that is, variables that have been measured on an ordinal, interval or ratio scale. The conducted research shows that the correlation coefficient can also be estimated for nominal data without classes of equal cardinality, which will be coded with real numbers. On the other hand, for data coded with complex numbers, it was not possible to unambiguously estimate the correlation coefficient.
	
	A common feature of the above-mentioned sets of real numbers, for which the linear correlation coefficient can be estimated, is that these sets can be sorted and, consequently, they can also be assigned ranks\footnote{For ranks, a correlation coefficient can be estimated. This is the so-called Spearman rank correlation coefficient.}. Thus, these are sets for which a linear order relation can be established. For sets containing complex numbers, linear order relations cannot be defined. The strongest relation possible in the set of complex numbers is the partial order relation.
	
	It can therefore be said that a necessary condition that should be met in order to be able to estimate the linear correlation coefficient is the possibility of establishing a linear order relation in both data sets representing the analyzed random variables.
	\subsection{Non-linear correlation}
	Due to the dependence of correlation on phase permutations in complex coding, an attempt was also made to search for a possible measure of non-linear correlation that would be invariant with respect to phase permutations in complex coding. For different permutations of phases in a complex variable $ V_1$, the correlation coefficient between a real variable $ V_2 $ and its non-linear approximation $ \widehat{V_2}=f(V_1) $ was investigated. For $ k $ different values of the variable $ V_1 $ containing $ m $ classes with equal cardinalities ($m\le{k}$), $ m! $ polynomials of degree $ k-1 $ were identified. In all cases, identical correlation coefficients were obtained. The obtained result is only partially satisfactory. Instead of the expected measure of linear correlation, this coefficient measures non-linear correlation.
	
	Regardless of this result, an unsolved problem beyond the scope of this article is the mathematical proof of the hypothesis that the linear correlation coefficient $ R(V_2,\widehat{V_2}) $ between the variable $ V_2 $ and the polynomial $ \widehat{V_2}=f(V_1) $ does not change with successive phase permutations.
	\subsection{Alternative solutions for classes with equal cardinalities}
	The analysis described in the article was conducted using artificial data, specially prepared for the purposes of this work. This was due to the fact that the author of the article did not find non-trivial nominal data sets that would contain classes of identical elements with equal cardinality. Thus, it cannot be excluded that the intuition that suggests that in large data sets it is unlikely that a given variable will contain two classes with equal cardinalities is true. It is even less likely that there could be more such classes.
	
	However, even if this is true, it is necessary to be able to deal with such potential problems. It seems that the methods used for pre-processing of missing or outlying data can be used here \cite{Maimon2010}\cite{Hand2001}\cite{Berthold2007}. For example, if the rejection of a single object causes two classes to no longer have equal cardinality, then the need for complex coding of the nominal variable also disappears.  On the other hand, if the analyzed sets have high cardinalities, then one can also reasonably hope that such an action will have a negligible effect on the statistics in the big dataset.
	
	The proposal presented here is not elegant. This proposal is a substitute. It can be used temporarily until further research provides a satisfactory solution to the problem presented here.
	\subsection{Possible simplification in nominal data coding}
	In subsection \ref{rangi}, the coding method is analogous to that used for ranking with tied ranks. A class of identical elements whose cardinality is equal to n is assigned the rank (\ref{eq16}). This approach was used in the calculations described in Section \ref{korelacje}. On the other hand, Subsection \ref{linear} noted that a linear transformation of a random variable does not change the direction of the vector representing its random component. Thanks to this, the coding method can be simplified for the purposes of correlation studies. In this case, it is enough to replace the ranks (\ref{eq16}) with cardinalities of the elements. The above change will give an unchanged value of the correlation coefficient, and at the same time slightly simplify the calculations performed during the coding procedure for the purpose of testing the strength of the correlation relationship. 
	\subsection{Possible further directions of research}
	The usefulness of the proposed method for assessing the linear correlation between nominal random variables coded with numbers requires further investigation. Regardless of the research carried out in this work, preliminary studies were carried out on the possibility of clustering numerically coded nominal variables into similarity classes (see Subsect. \ref{binary}) based on the correlation matrix. As a result, promising results have been obtained, which are a positive incentive for further work in this area.
	
	In subsection \ref{nonlin} it is noted that the linear correlation $R(V_2,\widehat{V_2})$ between the variable $V_2$ and the polynomial $\widehat{V_2}=f(V_1)$ does not change with successive phase permutations at complex coding of the nominal variable $ V_1 $. The above result is only partially satisfactory, because instead of the expected measure of linear correlation, it gives a measure of non-linear correlation. Regardless of this, it would be interesting to solve the theoretical problem of proving or rejecting the hypothesis that these correlation coefficients are invariant with respect to phase permutations in the complex coding of nominal data. 
	
	\section*{Acknowledgments}
	The author would like to express his gratitude to Dr. Leszek Rudak for valuable comments that helped to improve the content of the article.
	
	\bibliography{NomCorr}\label{bibliography}

\begin{thebibliography}{10}
\providecommand{\url}[1]{#1}
\csname url@samestyle\endcsname
\providecommand{\newblock}{\relax}
\providecommand{\bibinfo}[2]{#2}
\providecommand{\BIBentrySTDinterwordspacing}{\spaceskip=0pt\relax}
\providecommand{\BIBentryALTinterwordstretchfactor}{4}
\providecommand{\BIBentryALTinterwordspacing}{\spaceskip=\fontdimen2\font plus
\BIBentryALTinterwordstretchfactor\fontdimen3\font minus
  \fontdimen4\font\relax}
\providecommand{\BIBforeignlanguage}[2]{{%
\expandafter\ifx\csname l@#1\endcsname\relax
\typeout{** WARNING: IEEEtran.bst: No hyphenation pattern has been}%
\typeout{** loaded for the language `#1'. Using the pattern for}%
\typeout{** the default language instead.}%
\else
\language=\csname l@#1\endcsname
\fi
#2}}
\providecommand{\BIBdecl}{\relax}
\BIBdecl

\bibitem{Blalock1960}
H.~M. Blalock, \emph{{Social Statistics}}.\hskip 1em plus 0.5em minus
  0.4em\relax McGraw-Hill, 1960.

\bibitem{Francuz2007}
P.~Francuz and R.~Mackiewicz, \emph{Liczby nie wiedz{\k{a}}, sk{\k{a}}d
  pochodz{\k{a}}. Przewodnik po metodologii i statystyce nie tylko dla
  psycholog{\'o}w}.\hskip 1em plus 0.5em minus 0.4em\relax Lublin: Wydawnictwo
  KUL, 2007.

\bibitem{Gniazdowski2015}
\BIBentryALTinterwordspacing
Z.~Gniazdowski and M.~Grabowski, ``{Numerical Coding of Nominal Data},''
  \emph{Zeszyty Naukowe WWSI}, vol.~9, no.~12, pp. 53--61, 2015. [Online].
  Available: \url{http://doi.org/10.26348/znwwsi.12.53}
\BIBentrySTDinterwordspacing

\bibitem{Gniazdowski2013}
\BIBentryALTinterwordspacing
Z.~Gniazdowski, ``Geometric interpretation of a correlation,'' \emph{Zeszyty
  Naukowe WWSI}, vol.~7, no.~9, pp. 27--35, 2013. [Online]. Available:
  \url{http://doi.org/10.26348/znwwsi.9.27}
\BIBentrySTDinterwordspacing

\bibitem{Gniazdowski2011}
\BIBentryALTinterwordspacing
------, ``O relacjach i algorytmach,'' in \emph{Zbi{\'o}r wyk{\l}ad{\'o}w
  wszechnicy popo{\l}udniowej: Algorytmika i programowanie. Zastosowania
  informatyki}.\hskip 1em plus 0.5em minus 0.4em\relax Warszawska Wy{\.z}sza
  Szko{\l}a Informatyki, 2011, pp. 265--286. [Online]. Available:
  \url{http://akademickaseriawwsi.wwsi.edu.pl/ksiazki/5/O_relacjachi_algorytmach.pdf}
\BIBentrySTDinterwordspacing

\bibitem{Stevens1946}
\BIBentryALTinterwordspacing
S.~S. Stevens, ``On the theory of scales of measurement,'' \emph{Science}, vol.
  103, no. 2684, pp. 677--680, 1946. [Online]. Available:
  \url{https://psychology.okstate.edu/faculty/jgrice/psyc3120/Stevens_FourScales_1946.pdf}
\BIBentrySTDinterwordspacing

\bibitem{StatSoft2011}
\BIBentryALTinterwordspacing
StatSoft, ``Elektroniczny podręcznik statystyki,'' 2011. [Online]. Available:
  \url{https://www.statsoft.pl/textbook/stbasic.html}
\BIBentrySTDinterwordspacing

\bibitem{Mehta1983}
\BIBentryALTinterwordspacing
C.~R. Mehta and N.~R. Patel, ``A network algorithm for performing fisher's
  exact test in $r\times c$ contingency tables,'' \emph{Journal of the American
  Statistical Association}, vol.~78, no. 382, pp. 427--434, 1983. [Online].
  Available: \url{https://doi.org/10.1080/01621459.1983.10477989}
\BIBentrySTDinterwordspacing

\bibitem{Wilcoxon1945}
F.~Wilcoxon, ``Individual comparisons by ranking methods,'' \emph{Biometrics
  Bulletin}, vol.~1, no.~6, pp. 80--83, 1945.

\bibitem{Mok2014}
\BIBentryALTinterwordspacing
T.~Mok and H.~B. Iz, ``Vector regression introduced,'' \emph{Journal of
  geodetic science}, vol.~4, no.~1, pp. 57--64, 2014. [Online]. Available:
  \url{https://core.ac.uk/download/pdf/61032201.pdf}
\BIBentrySTDinterwordspacing

\bibitem{whuber}
\BIBentryALTinterwordspacing
whuber, ``Analysis with complex data, anything different?'' 2013. [Online].
  Available: \url{https://stats.stackexchange.com/q/66268}
\BIBentrySTDinterwordspacing

\bibitem{Dryja1982}
M.~Dryja, J.~Jankowska, and M.~Jankowski, \emph{Przegl{\k{a}}d metod i
  algorytm{\'o}w numerycznych, cz{\k{e}}{\'s}{\'c} 2}.\hskip 1em plus 0.5em
  minus 0.4em\relax Warszawa: Wydawnictwa Naukowo-Techniczne, 1982.

\bibitem{Kielbasinski1992}
A.~Kie{\l}basi{\'n}ski and H.~Schwetlick, \emph{Numeryczna algebra liniowa:
  wprowadzenie do oblicze{\'n} zautomatyzowanych}.\hskip 1em plus 0.5em minus
  0.4em\relax Wydawnictwa Naukowo-Techniczne, 1992.

\bibitem{Maimon2010}
\BIBentryALTinterwordspacing
O.~Maimon and L.~Rokach, Eds., \emph{Data mining and knowledge discovery
  handbook}.\hskip 1em plus 0.5em minus 0.4em\relax Springer, 2010. [Online].
  Available: \url{https://link.springer.com/content/pdf/10.1007/b107408.pdf}
\BIBentrySTDinterwordspacing

\bibitem{Hand2001}
D.~J. Hand, H.~Mannila, and P.~Smyth, \emph{Principles of data mining}.\hskip
  1em plus 0.5em minus 0.4em\relax MIT press, 2001.

\bibitem{Berthold2007}
M.~Berthold and D.~J. Hand, \emph{Intelligent data analysis. An
  introduction}.\hskip 1em plus 0.5em minus 0.4em\relax Springer, 2007.

\end{thebibliography}
	\bibliographystyle{IEEEtran}
	
\end{document}